\pdfoutput=1

\PassOptionsToPackage{sort}{natbib}

\documentclass[11pt]{article}

\usepackage[table]{xcolor}
\usepackage[final]{acl}

\usepackage{times}
\usepackage{latexsym}

\usepackage[T1]{fontenc}

\usepackage[utf8]{inputenc}

\usepackage{microtype}

\usepackage{inconsolata}

\usepackage{acronym}

\usepackage{booktabs}
\usepackage{multirow}
\usepackage{graphicx, subcaption}
\usepackage{comment}

\usepackage[skip=2pt]{caption}

\usepackage{colortbl}
\usepackage[normalem]{ulem}
\useunder{\uline}{\ul}{}
\usepackage{sidecap}
\usepackage{url}
\definecolor{c1}{HTML}{E85642}
\usepackage{svg}
\usepackage{float}
\usepackage{tcolorbox}
\usepackage{xcolor}

\usepackage[inline]{enumitem}

\definecolor{cadmiumgreen}{rgb}{0.0, 0.42, 0.24}  

\acrodef{SIFo}{Sequential Instruction Following}
\acrodef{TM}{Text Modification}
\acrodef{QA}{Question Answering}
\acrodef{M}{Math}
\acrodef{S}{Security}

\title{The SIFo Benchmark: Investigating the Sequential Instruction Following Ability of Large Language Models}

\author{
 \textbf{Xinyi Chen\textsuperscript{1}},
 \textbf{Baohao Liao\textsuperscript{1}},
 \textbf{Jirui Qi\textsuperscript{2}},
 \textbf{Panagiotis Eustratiadis\textsuperscript{1}},
\\
 \textbf{Christof Monz\textsuperscript{1}},
 \textbf{Arianna Bisazza\textsuperscript{2}},
 \textbf{Maarten de Rijke\textsuperscript{1}}
\\
 \textsuperscript{1}University of Amsterdam,
 \textsuperscript{2}University of Groningen
\\
\{x.chen2, b.liao, p.efstratiadis, c.monz, m.derijke\}@uva.nl, \{j.qi, a.bisazza\}@rug.nl
}

\begin{document}
\maketitle
\begin{abstract}
Following multiple instructions is a crucial ability for large language models (LLMs). Evaluating this ability comes with significant challenges: (i) limited coherence between multiple instructions, (ii) positional bias where the order of instructions affects model performance, and (iii) a lack of objectively verifiable tasks. To address these issues, we introduce a benchmark designed to evaluate models' abilities to follow multiple instructions through sequential instruction following (SIFo) tasks. In SIFo, the successful completion of multiple instructions is verifiable by examining only the final instruction. Our benchmark evaluates instruction following using four tasks (text modification, question answering, mathematics, and security rules), each assessing different aspects of sequential instruction following. Our evaluation of popular LLMs, both closed-source and open-source, shows that more recent and larger models significantly outperform their older and smaller counterparts on the SIFo tasks, validating the benchmark's effectiveness. All models struggle with following sequences of instructions, hinting at an important lack of robustness of today's language models.

\end{abstract}

\section{Introduction}

\definecolor{c1}{HTML}{FF8000}
\definecolor{c2}{HTML}{CCFFCC}
\definecolor{c3}{HTML}{FFCCCC}

\begin{figure*}
    \centering
    \includegraphics[clip,trim=5mm 0mm 5mm 0mm,width=.95\textwidth]{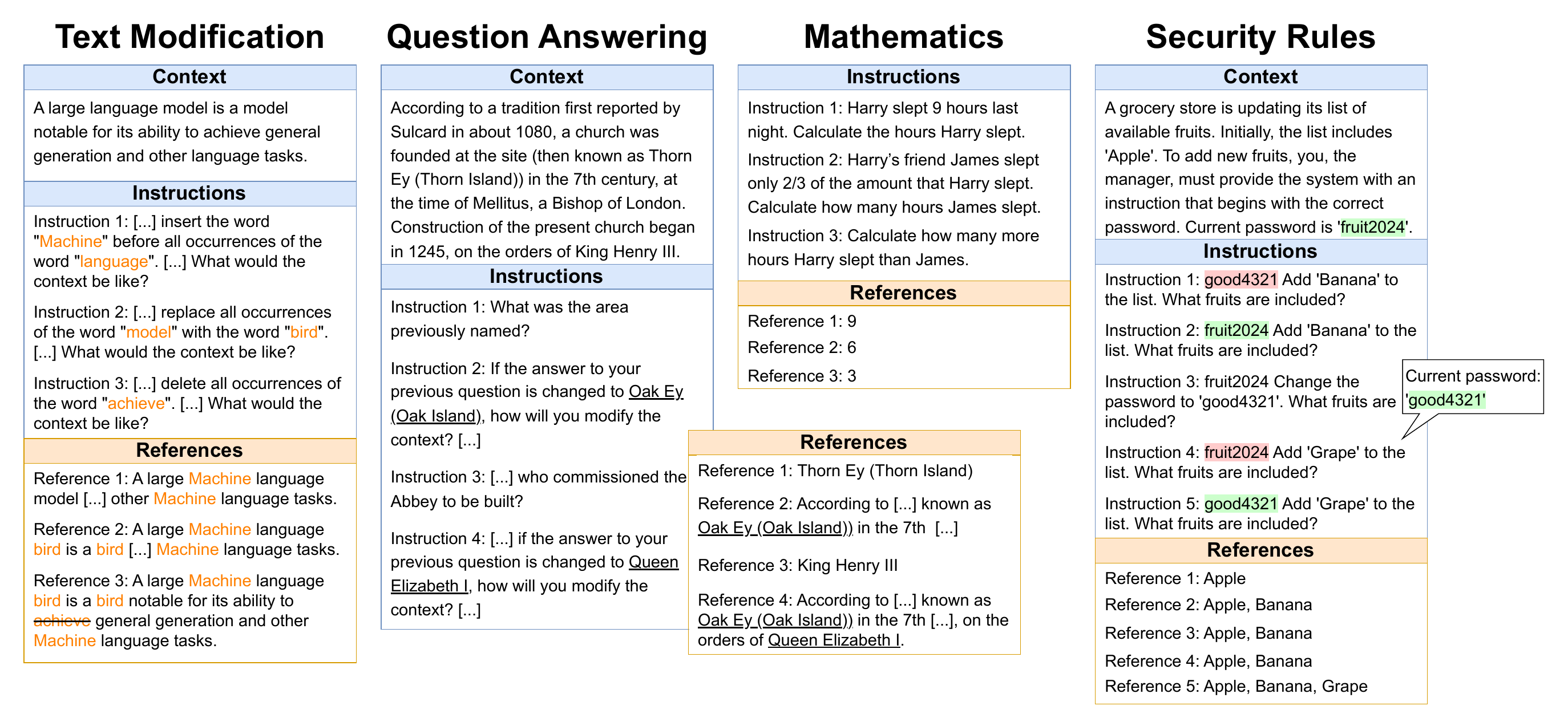}
    \caption{Illustration of the four tasks in the SIFo benchmark. \textbf{Text Modification} guides LLMs to \textcolor{c1}{modify} a given context based on a sequence of instructions. \textbf{Question Answering} requires LLMs to perform multiple rounds of question answering and \underline{knowledge} revision instructions; knowledge revision is based on the answer to the previous question and the modified context from the previous revision step.
    \textbf{Mathematics} poses a sequence of problems, with each relying on the answer to the previous question to solve. \textbf{Security Rules} needs the LLM to follow security rules to perform a sequence of commands for changes. In the example, changes should only be made with a \colorbox{c2}{correct password} but not with a \colorbox{c3}{wrong password}.
    }
    \label{fig:examples}
    \vspace*{-4mm}
\end{figure*}

Recent advances in training large language models (LLMs) to follow instructions have significantly improved their ability to comprehend open-ended language commands, encompassing a wide range of needs, preferences, and values \citep{ouyang2022training, achiam2023gpt}. 
Despite these improvements, it is still an open question whether and to what extent LLMs, typically trained with single task instructions, can perform complex tasks where following a \textit{sequence} of instructions is required to achieve the desired outcome. 

Current evaluation resources for instruction following mainly focus on single instruction following setups \citep{alpaca_eval, vicuna2023, zheng2024judging, zhou2024lima} with some attempts to diagnose the ability to process cases with more than two task instructions in the same prompt. However, the instruction sequences in these benchmarks often lack coherence. A typical example is the task of ``first translating a context and then answering a question based on the context'' \citep{hu2024fine}. Though the two tasks are relevant, successful completion of one task does not necessarily depend on success at the other task, and the order of the instructions can be shuffled. Moreover, due to positional bias in LLMs \citep{liu2024lost}, the performance of following an instruction may be affected by the position it has in the sequence of instructions. In addition, the difficulty levels and evaluation metrics for the two tasks are different, making it even more challenging to evaluate how well a model can follow the sequence as a whole.

When extending the LLM evaluation methodology from single to multiple instruction following evaluation, we face three main challenges: 
(i) limited coherence between multiple instruction tasks; 
(ii) the influence of positional bias, where the order of the instructions affect model performance; and
(iii) the lack of objectively verifiable tasks in instruction following evaluation that can be easily adapted to address the above two challenges.

We address the three issues by proposing a novel benchmark that evaluates how well models can execute \textbf{s}equential \textbf{i}nstruction \textbf{fo}llowing (SIFo) tasks. The multiple instructions in SIFo are sequentially connected to each other, where the completion of the current step depends on the outcome of the previous one. Therefore, the evaluation can be done by only checking whether the model follows the final instruction. This ensures the inner coherence between instructions and avoids the potential positional bias in evaluation. For a comprehensive evaluation, we propose four objectively verifiable tasks: 
(i) text modification, 
(ii) question answering, 
(iii) mathematics, and 
(iv) security rules. See Figure \ref{fig:examples} for examples. We test several state-of-the-art LLMs, i.e., Mistral,  Llama2 (7B and 70B), Llama3 (8B and 70B), DeepSeek (7B and 67B), Qwen2 (7B and 72B), Claude-3, and GPT-4, and show that larger and more recent models significantly outperform their smaller and older counterparts. Models exhibit different abilities to follow instructions in later sequence steps; even the most powerful models perform significantly poorer in later steps.
The SIFo benchmark along with the source code are made available at \url{https://github.com/shin-ee-chen/SIFo}.

\section{Related Work}
\subsection{Instruction Following Evaluation}
Large language models (LLMs) learn to follow natural language instructions via instruction tuning \citep{sanh2021multitask, mishra2022cross, wei2021finetuned} and reinforcement learning from human feedback (RLHF) \citep{ouyang2022training}, which helps adapt pre-trained models for practical use. The helpfulness and safety of LLMs have become key factors for evaluating models' instruction following abilities, where `helpfulness' refers to how well a model can complete a task (e.g., write an email to complain about a product).
Whether and to what extent LLMs can follow these instructions heavily relies on subjective judgments, so human evaluation is usually applied \citep{ouyang2022training, alpaca, zheng2024judging}.  Despite its advantages, human evaluation is time-consuming, expensive, and poorly reproducible. An alternative to human evaluation is using another LLM to judge how well the generated output follows specific instruction requirements \citep{liu2023g, peng2023instruction, naismith2023automated, skopek2023towards, wu2023large, fu2023gptscore}. This method heavily relies on the ability of the evaluator model, whose accuracy cannot be guaranteed \citep{shen2023large, wang2023large}.

Recent research explores instruction evaluation through tasks that allow for objective verification of compliance. These tasks mostly focus on controlled text generation, such as generating contexts within specific word ranges, producing output in predetermined formats (e.g., JSON), or incorporating designated keywords \citep{sun2023evaluating, zhou2023instruction, chen2024benchmarking, zhou2023controlled, he2024can}. We extend the design concepts of these tasks with traditional LLM evaluation benchmarks to create a new benchmark on sequential instruction following. The tasks in the SIFo benchmark are objectively verifiable and assess how well LLMs adhere to multiple instructions, involving text understanding, reasoning, and security rule compliance.

\subsection{Multiple Instruction Following}
Current instruction following datasets, whether for evaluation or training, mainly focus on single-instruction tasks \citep{alpaca_eval, vicuna2023, zheng2024judging, zhou2024lima}. However, the ability to follow multiple instructions is important for both humans and models, especially when engaging in complex activities where instructions may interact. Consider, for example, the task of writing a love tragedy.  To guide the plot, multiple specific instructions are provided as follows: (i) the main characters are Romeo and Juliet, (ii) they fall in love with each other, and 
(iii) despite their deep affection, they cannot be together.

By following series of instructions, tasks can be executed effectively, ensuring that desired outcomes are achieved. Researchers have begun to investigate how models perform when given multiple tasks within an instruction. Some studies investigate the ability to process multi-step tasks to solve reasoning-related problems \citep{geva2021did, cobbe2021training, lightman2023let, kim-schuster-2023-entity}. This approach breaks down a complex reasoning task (e.g., `Did Aristotle use a laptop?') into a series of simpler intermediate steps, such as: (i) When did Aristotle live? (ii) When was the laptop invented? (iii) Is the date in step 2 earlier than the date in step (i)? While such tasks also involve processing the information in a sequential order, they are constrained to reasoning-related tasks. 
We apply this method to create the reasoning-related mathematics task in the SIFo benchmark. Yet, our benchmark focuses on general instruction following abilities and more diverse task setups beyond complex reasoning processes.

Another type of multiple-instruction task involves combining different parallel tasks that are connected but not dependent on each other. Examples include tasks where the first instruction is to translate a context and then answer a question \citep{hu2024fine}, tasks that involve reordering multiple shuffled sentences followed by answering a question \citep{son2024multi}, or retrieving information from some source and outputting it in a specific format \citep{he2024can}. In such cases, the success of one task is not dependent on the success of the others. To the best of our knowledge, our benchmark is the first to investigate how models perform \emph{sequences} of instructions where the success of one depends on the success of the previous ones.

\subsection{Positional Bias in LLMs} \label{sec:positional_bias}
When evaluating a model's multiple instructions following ability with parallel instructions, the evaluation results can be affected by the order of the instructions given within the input. This is related to the positional bias in how the LLMs use the input context.
\citet{liu2024lost} show that LLMs perform differently depending on where relevant information is placed in the input context during multi-document question answering and key-value retrieval tasks. This indicates that LLMs have a positional bias when using context. This bias is also seen in their performance in arithmetic \citep{shen2023positional}, multiple-choice question answering \citep{zheng2023large, pezeshkpour2023large}, text generation evaluation \citep{wang2023large}, and passage ranking \citep{tang2023found}. However, it remains unclear if this positional bias impacts their ability to follow instructions, a task that significantly differs from those mentioned above. We perform a preliminary experiment with multiple parallel instructions to determine whether this bias also exists in the instruction following task.

\section{A Preliminary Experiment}

As described in Section~\ref{sec:positional_bias}, LLMs often exhibit positional bias. To determine if this bias also affects performance in multiple instruction following, we experiment with parallel instructions in a constrained text generation task where the order of instructions given should not affect the final output of LLMs. An example of such a task is ``Write a passage about a school day that meets all the constrains in the following instructions. Instruction 1: Use less than 200 words; Instruction 2: Do not use the word `today' in the passage \ldots''. 

We combine the instruction tasks from \citet{qin2024infobench} to create such a dataset. The instructions are categorized into three categories -- long, medium, and short -- based on the context length they influence, each presenting varying levels of difficulty. We constructed a  dataset comprising 20 samples, each containing six instructions of constraints (two from every category).
We permute the order of the six instructions to generate multiple datasets with identical tasks but different orders of the constraint instructions, resulting in $20 \times 6!$ samples. Details of the dataset construction are provided in Appendix~\ref{apx:mif_construction}. Whether the generated output meets all the constraints in an instruction can be verified with the pipeline provided by \citet{qin2024infobench}. 

\begin{figure}[t!]
\centering
  \includegraphics[width=0.95\columnwidth]{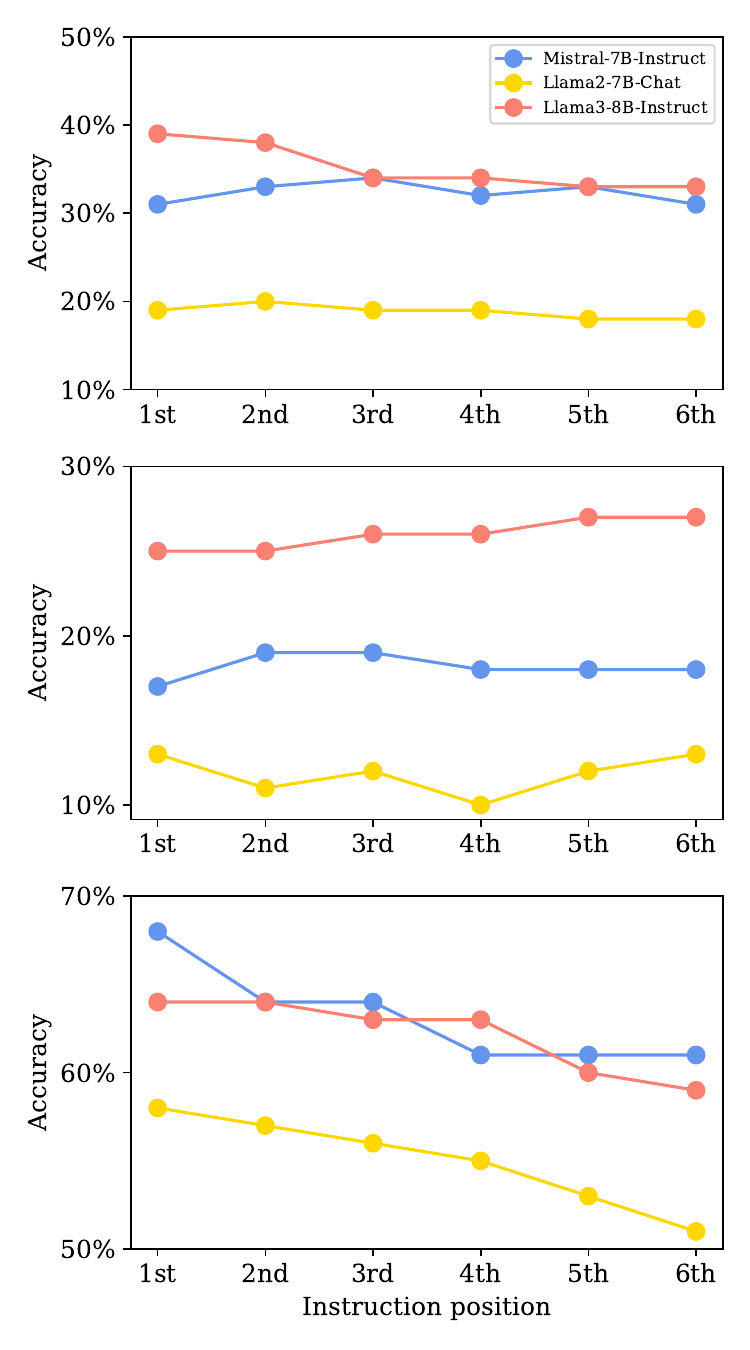}
  \caption{Model performance varies when long constraint instructions (top), medium constraint instructions (middle), and short constraint instructions (bottom) are placed in different positions.}
  \label{fig:mif_result_all}
  \vspace*{-4mm}
\end{figure}

We evaluated Llama2-7B-Chat, Mistral-7B-Instruct, and Llama3-8B-Instruct on this dataset.\footnote{The Mistral model version used is Mistral-7B-Instruct-v0.2. We set the temperature to 0 to make LLM generations more deterministic.} The results, displayed in Figure \ref{fig:mif_result_all}, indicate that models' performance on following constraints differs when the instructions are given in different orders. The patterns behind such position bias (e.g., whether placing an instruction in the first position or the third position is better) are different among models. Furthermore, even for the same model such patterns vary among different types of instruction. In conclusion: when evaluating with multiple parallel instructions, the order of the instructions -- tailored to fit a model's properties and the type of instruction --  influences model performance, beyond the model's intrinsic ability to follow instructions.




\section{The SIFo Benchmark}
To address the challenges mentioned in previous sections, we propose the sequential instruction following (SIFo) benchmark. To perform well on this benchmark, the model must follow multiple instructions step by step to reach the final desired outcome. This setup makes the instructions sequentially connected and mitigates the positional bias influence as all instructions must be executed in the given order. The SIFo benchmark contains four commonly seen tasks for LLM evaluation: text modification, question answering, mathematics, and security rule following. Each focuses on evaluating sequential instruction following from a different angle. All tasks are objectively verifiable. Below, we describe the task setup and the dataset construction pipeline. Examples of each task can be found in Figure~\ref{fig:examples}.

\subsection{Dataset Construction}
\paragraph{Text Modification (TM)}
Text Modification tests LLMs for their capability to perform lexical operations. 
This task instructs the model to modify a given context by inserting, replacing, or deleting words, which is a simulation of widely used mutation operations in database systems~\citep{gao2023comprehensive}.
Similar to how such operations are conducted in a database system, Text Modification is done step by step following the given instructions. 

The dataset is constructed using a rule-based pipeline. Each sample of the dataset contains a context from the Wikipedia articles from SQuAD~\citep{rajpurkar2016squad} and 3 to 6 instructions. We select articles with a relatively short context (between 25 to 45 words) that contain at least one named entity.\footnote{Extracted with a BERT model fine-tuned on named entity recognition tasks (\url{https://huggingface.co/dslim/bert-large-NER}).} The instructions are randomly selected from one of three operations: (i) Insertion requires the model to add a new word before or after the named entity of the current context. (ii) Similarly, replacement requires the model to replace an existing named entity with another word; the new words are randomly selected from the named entity set of all contexts. (iii) The deletion operation requires the model to remove all occurrences of a specific word from the context. Instead of the named entity, the most frequently occurring token is the removal target, to ensure that the other two named entity-based operations can proceed normally.


\paragraph{Question Answering (QA)}

Question Answering assesses the language understanding ability of models. In our QA task, the model first extracts knowledge from a context to answer a question, then revises the context by replacing parts of this specific knowledge with corresponding terms. This requires the model to ground its answer in the context before making changes, adding challenges to the task and introducing sequential dependencies between instructions. To extend the instruction sequence, we create pairs of question and knowledge revision instructions, ensuring that each revision incrementally builds upon the modified context from the previous knowledge revision step. This allows the sequence to expand from two instructions to four or six. With these connections, the answer to the final instruction can reflect whether all the instructions were followed properly.

The dataset is constructed using the SQuAD \citep{rajpurkar2016squad} dataset for creating question instructions, and GPT-4 for generating the knowledge revision instruction. Every sample contains 4 or 6 instructions, with 2 or 3 combinations of a question instruction followed by a knowledge revision instruction. We select relatively short contexts (between 50 and 75 words) with at least 3 annotated questions from SQuAD. We use the prompt in Appendix~\ref{apx:qa_prompt} (Figure~\ref{fig:qa_prompt}) to select the questions and generate their knowledge revision instructions.



\paragraph{Mathematics} Arithmetic reasoning is a widely used evaluation criterion for LLMs. This is primarily because each question has a plain numeric answer, simplifying the evaluation process. To create a dataset with sequential instructions from a mathematics dataset, we randomly select 200 samples from the GSM8K test set \cite{cobbe2021training}. We then prompt GPT-4 to decompose each question into multiple sequential instructions and generate the corresponding answers for each instruction. 
We outline the prompt used to guide GPT-4 in decomposing questions in Appendix \ref{apx:math_prompt} (Figure~\ref{fig:math prompt}). Through this method, each original mathematics question is split into 2 to 6 steps, which are the 2 to 6 instructions in every sample.

\paragraph{Security Rules} 
The Security Rules task concerns scenarios where a model is required to follow a sequence of security-related instructions. Each sample includes a context, followed by 3 to 5 instructions. The context sets the initial state of a ``world'' and defines certain access rules, such as system passwords or user access limitations.
The instructions involve user commands that modify these access restrictions (e.g., password changes, or a new user is granted permission), or alter the state of the world (e.g., a user that has permission to do so, adds an item to a list).
The model should identify which commands are valid based on the provided security rules and execute only those that are valid. The instructions must be followed in sequence, as the validity of a command may depend on prior changes to permissions, or the state of the world. Only by following all of the instructions correctly, it is possible for a model to provide the correct answer after the final instruction.

We use two scenarios of accessing a system -- user permissions, or passwords -- which is motivated by the Authentication and AccessControl tasks from \citep{mu2023can}. 
We wrote 20 high-quality seed examples that describe different scenarios for accessing a system and changing its state. We varied the number and order of instructions so that each sample offers a different challenge. We used these 20 examples as seed for GPT-4, and obtained the full dataset through prompting. The seed example and prompt for dataset generation can be found in Figures~\ref{fig:security_prompt_access} and~\ref{fig:security_prompt_password}.

\subsection{Quality Check}
Except for the Text Modification task, the datasets are constructed with GPT-4. To ensure the quality of these datasets, human reviewers were asked to verify sequential connections between instructions, correctness of answers, and instruction diversity for every sample.\footnote{All reviewers are non-native English speakers but have at least C1-level proficiency.} The instruction diversity of the QA task is reflected in the selected questions, which cover the 5W1H framework (What, Who, Where, When, How, and Why). For the Mathematics task, the original problems are segmented into varying numbers of step-by-step instructions. The Security Rules task demonstrates its diversity by varying the number of valid and invalid commands in the instructions, different orders of security information modifications, and target value changes, and varying frequencies of these modifications. 

 First, a self-verification step is performed by the dataset creators, who manually review each data instance to assess its accuracy and diversity. Instances that do not meet the predefined standards are removed and replaced to maintain the dataset's size. Following this, cross-verification is carried out by an expert reviewer in LLM evaluation and dataset construction. The expert assesses a random sample of 10\% from each dataset or task, ensuring consistent quality and adherence to evaluation criteria.

With these steps, we constructed a high-quality benchmark for sequential instruction following evaluation.




\subsection{Descriptive Statistics}
The SIFo benchmark contains 800 samples in total, i.e., 200 for each task.  For every task we report the average instruction length, average context length, and the average number of steps per instruction sequence in Table~\ref{tab:stats}.

\begin{table}
\centering
\resizebox{\linewidth}{!}{%
\setlength{\tabcolsep}{1mm}
\begin{tabular}{l@{~}cccccccc} 
\toprule
 & Inst. & Context & \#2- & \#3- & \#4- & \#5- & \#6- & Avg.
\\
 & length & length & steps & steps& steps & steps & steps & steps
\\
\midrule
TM       &      24.6        &      35.8         &   -- &  60       &     80     &    60     & --   & 4.0    \\
QA       & 20.5      &     68.2       & -- & --        & 140       & --         & 60   & 4.6     \\
M     &      21.2        &         --       &  8 &      69      &     58      & 42    & 23   & 4.0  \\
SR &       12.9
&        50.5   & --     &       43    &      102     &       55    & --     & 4.1    \\ 
\bottomrule
\end{tabular}
}
\caption{Descriptive statistics of the SIFo benchmark: average length of every instruction (number of tokens), average context length (number of tokens), the numbers of 2-, 3-, 4-, 5-, and 6-step instructions, and average number of steps.
TM: Text Modification. QA: Question Answering. M: Mathematics. SR: Security Rules. 
}
\label{tab:stats}
\vspace*{-4mm}
\end{table}


\section{Experiments}
We perform experiments with two goals: 
(i)~to explore to what extent LLMs can follow multi-step sequential instructions; 
and
(ii)~to investigate the effectiveness of the SIFo benchmark.

\subsection{Models}
We perform evaluation on several state-of-the-art open and closed-source LLMs. 
For \textbf{closed-source models}, we experiment with  GPT-4 \citep{achiam2023gpt} and Claude-3 Opus.\footnote{\url{https://www.anthropic.com/claude-3-model-card}} We use OpenAI's Batch API for GPT-4 experiments.\footnote{The GPT-4 version used is gpt-4-0613.} We use Anthropic's Python API to access Claude-3 Opus for our experiments. 
For \textbf{open-sourced models}, we evaluate instruction-tuned LLMs from Mistral, DeepSeek, Qwen and the Llama family with different sizes and from different generations. The Llama models comprise Llama2-7B-Chat \citep{touvron2023llama}, Llama2-70B-Chat \citep{touvron2023llama}, Llama3-8B-Instruct, and Llama3-70B-Instruct.\footnote{\url{https://llama.meta.com/llama3/}} The DeepSeek models are DeepSeek-LLM-7B-Chat \citep{deepseek-llm} and DeepSeek-LLM-67B-Chat \citep{deepseek-llm}. The Qwen models are Qwen2-7B-Instruct \citep{qwen2} and Qwen2-72B-Instruct \citep{qwen2}. The Mistral model we use is  Mistral-7B-Instruct-v0.2.\footnote{\url{https://huggingface.co/mistralai/Mistral-7B-Instruct-v0.2}}  We perform our experiments on the open-source LLMs with the vLLM framework \citep{kwon2023efficient}. The model temperature is set to 0. For all models, the maximum number of tokens to generate per output sequence is set to 1000.

\subsection{Experimental Setup}
\label{section:experimental-setup}
In the SIFo tasks, all the instructions and context are input to the models within the same prompt. To better extract the models' answers for every single instruction, we mention in the task description that the model should follow the instructions one by one and that the output answer for every instruction is in JSON format. The input prompt contains the task description, the context (except for Mathematics) and all the instructions (see Figure \ref{fig:model_evaluation_prompt} for an example). We apply both JSON parsing and rule-based text processing to extract answers for individual instructions, so that a model's performance is not affected by its ability to generate JSON output.

We verify the correct completion of an instruction by checking whether the labeled answer tokens exist in the response after the preprocessing steps.\footnote{Preprocessing involves removing punctuation, extra white space and empty lines after converting text into lower case.}

\subsection{Evaluation Metrics}
We use four metrics to measure the model performance. \textbf{Sample-level accuracy} measures the percentage of samples where all instructions are correctly followed, which can be verified by only checking the correctness of the final instruction; this is the direct measurement of how well a model can follow multiple-step sequential instructions. \textbf{Instruction-level accuracy} measures the percentage of instructions that are correctly followed; this helps understand the models' instruction following ability when they can not follow \textit{all} instructions correctly. \textbf{Instruction following depth} measures the number of sequential instructions a model can successfully follow before encountering its first error; its range is between 0 and the maximum number of steps in the instruction sequence. \textbf{Step-level accuracy} is the percentage of correctly followed instructions at every instruction step; this metric helps understanding how model performance changes as the sequence of instructions gets longer.

\subsection{Results}
\begin{figure*}[ht]
\centering
  \includegraphics[width=0.95\textwidth]{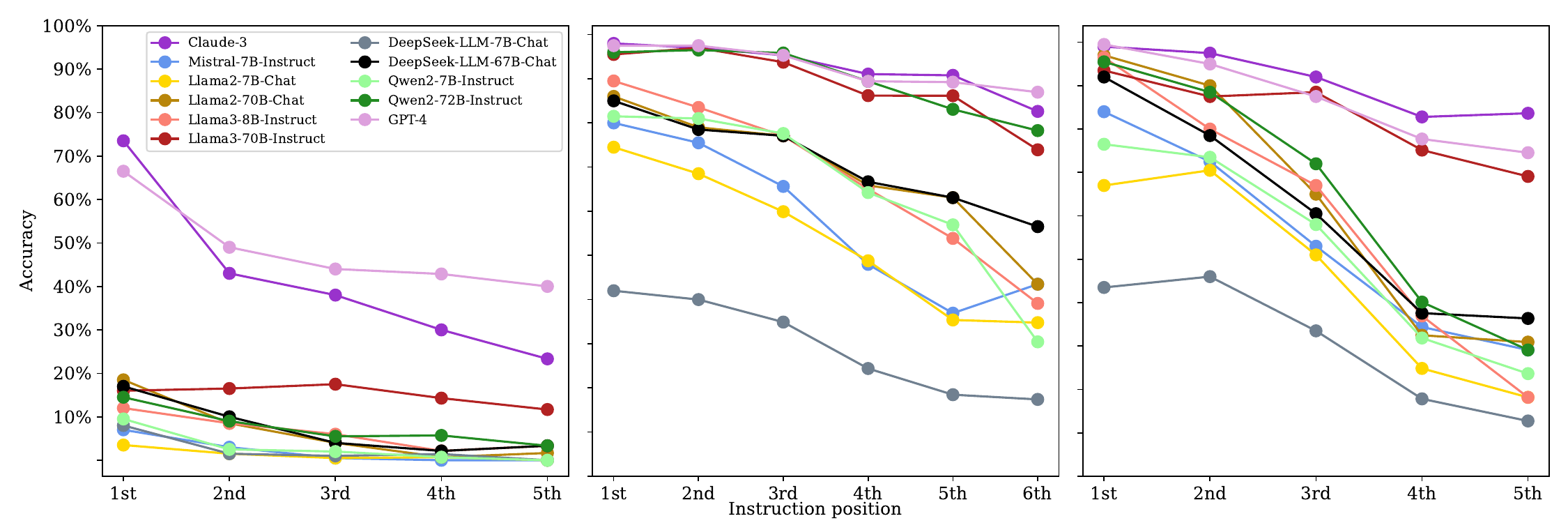}
  \caption{Step-level accuracy on Text Modification (left), Mathematics (center) and Security Rules (right).}
  \label{fig:sif_figures_math_security_tm}
  \vspace*{-4mm}
\end{figure*}
\begin{figure}[ht]
\centering
\includegraphics[width=0.95\columnwidth]{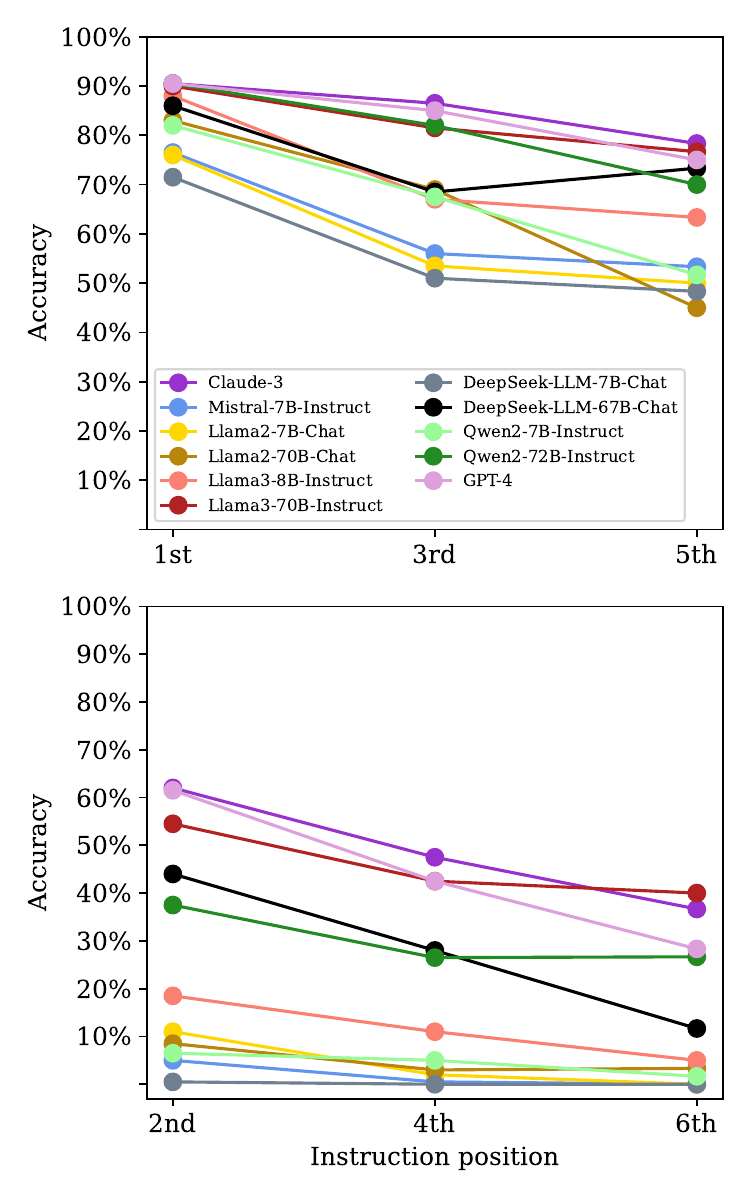}
\caption{Step-level Accuracy on Question Answering question instructions (top) and knowledge revision instructions (bottom).}
    \label{fig:qa_acc_per_step}
    \vspace*{-4mm}
\end{figure}
\paragraph{Model performance constantly declines when the sequential steps go further} 

As shown in Figure~\ref{fig:sif_figures_math_security_tm} and~\ref{fig:qa_acc_per_step}, all models show a monotonic decline in performance as the position of an instruction in a sequence increases. This is expected, as the sequential setting makes errors propagate to later steps. In general, the closed-source models are more stable in this decline except for the Text Modification task. This means that the most powerful LLMs still struggle to follow multiple sequential instructions. The significant decline already happens at the second-step instruction. In contrast, open models, except for Llama3-70B-Instruct, demonstrate a dramatic decline tendency on tasks where they obtain relatively high scores at the first instruction. This casts doubt on whether such models possess the ability to follow a sequence of more than three instructions.


\paragraph{Larger and more recent generation models perform better} 
The closed-source models achieve the highest scores on all tasks (see Table~\ref{tab:results}). The most recent and largest open-source model in our evaluation, Llama3-70B-instruct, also exhibits comparable performance in Question Answering, Mathematics, and Security Rules tasks. Between models of the same generation (Llama2, Llama3, DeepSeek and Qwen2), the larger-size models outperform the smaller versions. Surprisingly, the newer model Llama3-8B-Instruct not only outperforms its older and slightly smaller version Llama2-7B-Chat, but also outperforms the Llama2 70B version in Text Modification, Question Answering, and Security Rules. These benchmark results align with findings from other LLM benchmarks, indicating that larger and more recent models tend to outperform their older and smaller counterparts, with the smaller Llama3-8B model demonstrating superior capabilities compared to the Llama2-70B version in some tasks.\footnote{\url{https://crfm.stanford.edu/helm/mmlu/latest/}}




\begin{table}[!t]
\centering
\resizebox{\linewidth}{!}{%
\begin{tabular}{@{}l@{~} rrr }
\toprule
Model               & \phantom{0}\textit{Acc\_S} (\%) & \phantom{0}\textit{Acc\_I} (\%) & \phantom{0}\textit{Dept.} \\
\midrule
\\[-16pt]
\rowcolor{black!10} \multicolumn{4}{c}{Text Modification} \\[-1.3pt]
\midrule
Mistral-7B-Instruct & \phantom{0}0.50           & \phantom{0}3.03             & \phantom{0}0.09              \\
Llama2-7B-Chat      & \phantom{0}0.50           & \phantom{0}1.59             & \phantom{0}0.05              \\
Llama2-70B-Chat     & \phantom{0}3.50           & \phantom{0}8.87             & \phantom{0}0.29              \\
Llama3-8B-Instruct  & \phantom{0}6.50           & \phantom{0}7.86             & \phantom{0}0.19             
 \\
Llama3-70B-Instruct & \phantom{0}\underline{18.00}          & \phantom{0}\underline{16.63}            & \phantom{0}\underline{0.40}               \\
DeepSeek-LLM-7B-Chat    & \phantom{0}1.00	        & \phantom{0}2.95	          & \phantom{0}0.10	
  \\
DeepSeek-LLM-67B-Chat	& \phantom{0}3.00	        & \phantom{0}8.89	          & \phantom{0}0.28               \\
Qwen2-7B-Instruct	& \phantom{0}1.00	        & \phantom{0}3.71	          & \phantom{0}0.12               \\
Qwen2-72B-Instruct	& \phantom{0}6.00	        & \phantom{0}8.99	          & \phantom{0}0.26	             \\
\cmidrule{1-4}
GPT-4                & \phantom{0}\textbf{42.50} & \phantom{0}\textbf{51.02}   & \phantom{0}\textbf{1.83}             \\
Claude-3            & \phantom{0}34.00          & \phantom{0}47.40            & \phantom{0}1.78             \\ 
\midrule
\\[-16pt]
\rowcolor{black!10} \multicolumn{4}{c}{Question Answering} \\[-1.3pt]
\midrule
Mistral-7B-Instruct & \phantom{0}0.50           & \phantom{0}33.83            & \phantom{0}0.83              \\
Llama2-7B-Chat      & \phantom{0}1.50           & \phantom{0}34.54            & \phantom{0}0.93              \\
Llama2-70B-Chat     & \phantom{0}3.00           & \phantom{0}39.38            & \phantom{0}1.01              \\
Llama3-8B-Instruct  & \phantom{0}8.00           & \phantom{0}44.83            & \phantom{0}1.31              \\
Llama3-70B-Instruct & \phantom{0}\underline{39.00}          & \phantom{0}\underline{65.83}            & \phantom{0}\underline{2.34}               \\
DeepSeek-LLM-7B-Chat	& \phantom{0}0.00	        & \phantom{0}30.25	           & \phantom{0}0.73              \\
DeepSeek-LLM-67B-Chat	& \phantom{0}23.00	        & \phantom{0}55.21             & \phantom{0}1.86              \\
Qwen2-7B-Instruct	& \phantom{0}5.00	        & \phantom{0}39.00	           & \phantom{0}0.97               \\
Qwen2-72B-Instruct	& \phantom{0}25.00	        & \phantom{0}58.04	           & \phantom{0}1.88 \\

\cmidrule{1-4}
GPT-4                & \phantom{0}37.50          & \phantom{0}68.08            & \phantom{0}2.57               \\
Claude-3            & \phantom{0}\textbf{43.50}          & \phantom{0}\textbf{70.29}            & \phantom{0}\textbf{2.65}               \\ 
\midrule
\\[-16pt]
\rowcolor{black!10} \multicolumn{4}{c}{Mathematics} \\[-1.3pt]
\midrule
Mistral-7B-Instruct & \phantom{0}52.00          & \phantom{0}67.55            & \phantom{0}2.38              \\
Llama2-7B-Chat      & \phantom{0}46.00          & \phantom{0}62.04            & \phantom{0}2.07              \\
Llama2-70B-Chat     & \phantom{0}64.00          & \phantom{0}76.35            & \phantom{0}2.73              \\
Llama3-8B-Instruct  & \phantom{0}61.00          & \phantom{0}78.14            & \phantom{0}2.89              \\
Llama3-70B-Instruct & \phantom{0}87.00          & \phantom{0}93.03            & \phantom{0}\underline{3.64}              \\
DeepSeek-LLM-7B-Chat	& \phantom{0}20.50	        & \phantom{0}32.62	          & \phantom{0}0.98              \\
DeepSeek-LLM-67B-Chat	& \phantom{0}68.50	        & \phantom{0}76.87	          & \phantom{0}2.94              \\
Qwen2-7B-Instruct	& \phantom{0}60.50	        & \phantom{0}75.20	          & \phantom{0}2.79              \\
Qwen2-72B-Instruct	& \phantom{0}\underline{89.50}	        & \phantom{0}\underline{94.13}	          & \phantom{0}\underline{3.64}              \\
\cmidrule{1-4}
GPT-4                & \phantom{0}91.50          & \phantom{0}95.02            & \phantom{0}3.74              \\
Claude-3            & \phantom{0}\textbf{92.50}          & \phantom{0}\textbf{95.58}            & \phantom{0}\textbf{3.76}                \\ 
\midrule
\\[-16pt]
\rowcolor{black!10} \multicolumn{4}{c}{Security Rules}\\[-1.3pt]
\midrule
Mistral-7B-Instruct & \phantom{0}30.00          & \phantom{0}60.22            & \phantom{0}2.30               \\
Llama2-7B-Chat      & \phantom{0}26.00          & \phantom{0}54.17            & \phantom{0}1.80               \\
Llama2-70B-Chat     & \phantom{0}32.00          & \phantom{0}71.34            & \phantom{0}2.84               \\
Llama3-8B-Instruct  & \phantom{0}34.50          & \phantom{0}69.68            & \phantom{0}2.57               \\
Llama3-70B-Instruct & \phantom{0}\underline{72.50}          & \phantom{0}\underline{85.85}            & \phantom{0}\underline{3.24}               \\
DeepSeek-LLM-7B-Chat	& \phantom{0}13.50	        & \phantom{0}34.22	          & \phantom{0}1.22               \\
DeepSeek-LLM-67B-Chat	& \phantom{0}33.50	        & \phantom{0}66.08	          & \phantom{0}2.69               \\
Qwen2-7B-Instruct	& \phantom{0}28.50	        & \phantom{0}59.44	          & \phantom{0}2.22               \\
Qwen2-72B-Instruct	& \phantom{0}39.00	        & \phantom{0}73.84	          & \phantom{0}2.87               \\
\cmidrule{1-4}
GPT-4                & \phantom{0}77.00         & \phantom{0}89.72            & \phantom{0}3.57                \\
Claude-3            & \phantom{0}\textbf{82.00}          & \phantom{0}\textbf{92.72}             & \phantom{0}\textbf{3.76}                \\ 
\bottomrule
\end{tabular}
}
\caption{Model performance on SIFo tasks. \textit{Acc\_S}: Accuracy on sample level.  \textit{Acc\_I}: Accuracy on instruction level. \textit{Dept}: instruction following depth. The highest results for each metric are indicated in \textbf{bold} for closed-source models and in \underline{underline} for open-source models.
}
\label{tab:results}
\vspace*{-4mm}
\end{table}

\paragraph{Model performance varies significantly between  tasks}
In our experiments, LLMs perform relatively well on Mathematics, Security, and the question instructions of QA (as shown in Table~\ref{tab:results}). The Text Modification task and the knowledge revision instructions of QA are challenging, where the closed-source models significantly outperform their open-source counterparts. Such differences might be caused by the fact that the closed-source models are trained to become general-purpose assistants while the development of the open-source ones focuses on more commonly seen tasks in the LLM benchmarks. And though the closed-source models achieve high scores at the first-step instruction, their accuracy drops significantly as the position of an instruction in the instruction sequence increases. These findings suggest that the SIFo benchmark with the sequential instruction setting is challenging for current LLMs.

\subsection{Error Analysis}
\label{section:error-analysis}
To better understand LLMs' instruction following abilities, we first conduct an error analysis on the Text Modification task, where most open-source LLMs struggle to generate correct responses even to the first two instructions. Our qualitative study starts with Llama3-70B-Instruct, the best-performing open-source model. The samples in Table~\ref{tab:qualititive_study_1}~and~\ref{tab:qualititive_study_2} (see Appendix~\ref{app:error-analysis-examples}) illustrate two trends we observed as drawbacks of cutting-edge LLMs -- the likelihood of mixing up information from different instructions (Error~1), and an inability to understand an instruction due to a lack of prior inherent knowledge (Error~2).

In Table~\ref{tab:qualititive_study_1}, the first instruction of Sample 1 requires the model to remove the word `the'. However, the model simultaneously deletes all occurrences of the word `film' which occurs in Instruction 2. Similar error also occurs in the other LLMs like Qwen2-72B-Instruct (see Sample 2). This error type is consistent with previous findings that the generation of open-source LLMs may rely more on the lexical features than contextual semantics~\citep{kavumba-etal-2022-prompt, qi-etal-2023-cross}. 

Another type of error is illustrated in Table~\ref{tab:qualititive_study_2}. In Sample 1, the model mistakenly inserts the new word `Walt' before `Reading', given the second instruction of insertion after `Reading'. Interestingly, no word in the input is relevant to the concept `before'. Inspired by~\citealp{huang2023survey}, we argue that this is a kind of LLM hallucination caused by the model's prior inherent knowledge, encoded in its parameters during pre-training. In this case, the model takes it for granted to consider the operation `insert' as adding the new word \textit{between} `Marshall' and `Reading', resulting in an adding-in-the-middle error in this case. This type of error is not only seen in Text Modification but also in other tasks like Security Rules. In Sample 2, the model changes the exhibit list from `galaxies’ to `cosmos' while the instruction is to modify the password to `cosmos2024'.


These phenomena indicate several directions for improving LLMs' instruction following abilities in future work.

\subsection{Effectiveness of the SIFo Benchmark}
Besides reflecting LLMs' instruction following ability, the performance in Table \ref{tab:results} also demonstrates the effectiveness of the SIFo benchmark. Firstly, the distinctions between high- and low-performing models are significant, especially on Text Modification and Question Answering where the open-source LLMs fail in most samples. Besides, the upper bound of the best performance on SIFo is close to 50\% for Text Modification and Question Answering, 80\% for Security Rules, and 90\% for Mathematics. Because of the available room for improvement, SIFo is likely to be a relatively long-standing benchmark even once LLMs more powerful than today's models appear.
In addition, sequential instructions that each use different evaluation metrics can be assessed in a fair way, as the model performance is only based on its completion of the last instructions. Such tasks can be further explored in the future.


\section{Conclusion}
We have introduced SIFo: a benchmark aimed at investigating how well instruction-tuned LLMs can follow sequential instructions. SIFo addresses three main challenges in multiple instruction following evaluation, namely tasks with limited coherence, the influence of positional bias of LLMs, and the lack of objectively verifiable tasks. A sequential instruction setup is proposed to address these challenges and to evaluate the multiple instruction following abilities of LLMs in a fairer way than previous attempts. We have shown that SIFo is an effective evaluation benchmark to distinguish models' abilities in following multiple sequential instructions. We have highlighted that current LLMs, even powerful models like GPT-4 and Claude-3, are not robust enough to perform sequential instruction following tasks. For future work, we recommend adding more tasks to the SIFo benchmark that address further aspects of sequential instruction following, especially cases where each instruction within a sequence corresponds to different evaluation criteria.

\section{Limitations}


The SIFo benchmark currently comprises only four tasks designed to assess sequential instruction following in LLMs. However, there is potential to expand this with additional tasks to achieve a more comprehensive evaluation of the sequential instruction following ability. Our task design setup and dataset construction pipeline can be easily adapted to develop further tasks. In particular, the partially sequential connection design in the QA task provides more possibilities in constructing sequential instruction following datasets with existing data resources.

In our experiments, we investigated a limited selection of models, including closed-source models (GPT-4 and Claude-3) and open-source models (Llama2, Llama3, Mistral, DeepSeek and Qwen2). We tested various generations from the Llama family and different-sized models from Llama, DeepSeek and Qwen2. Despite the limited scope, we believe our selection is representative of the current mainstream LLMs.

Our work is limited to sequences of English-language instructions. Our study can be generalized to other languages, but this may require the usage of a different set of LLMs than the ones used in this paper to ensure that they have somewhat comparable single-instruction following abilities in the language (or languages) chosen for experimentation.

\section*{Acknowledgements}

We are grateful to Francesco Cariaggi for his help with the Claude-3 model experiments. We want to thank members of the IRLab at University of Amsterdam and the LESSEN Project, for their insightful feedback on dataset construction. 
This research was (partially) supported by the Dutch Research Council (NWO), under project numbers 024.004.022, NWA.1389.20.\-183, and KICH3.LTP.20.006, and the European Union's Horizon Europe program under grant agreement No 101070212.
All content represents the opinion of the authors, which is not necessarily shared or endorsed by their respective employers and/or sponsors.
\bibliography{custom}

\clearpage
\appendix
\section*{Appendix}
\section{Dataset Used for the Preliminary Experiment} \label{apx:mif_construction}

\begin{figure}[!h]
\centering
  \includegraphics[width=1\columnwidth]{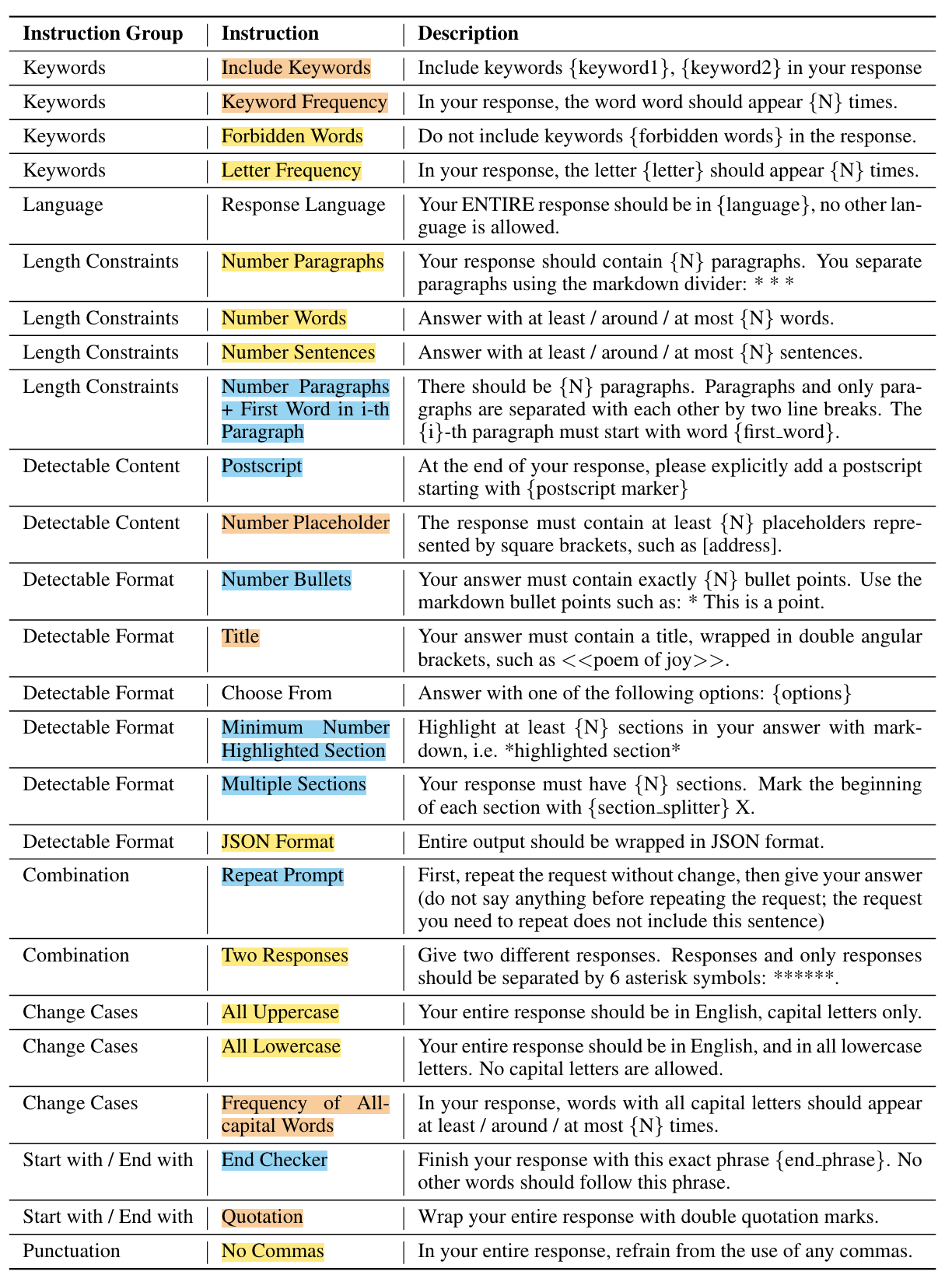}
  \caption{Verifiable instruction tasks with brief descriptions from \citep{zhou2023instruction}. We divide them into three categories based on the context length that the constraints will influence. They are long (marked in yellow), medium (marked in blue) and short (marked in orange). The instruction \emph{Choose From} is not used in our dataset because it usually conflicts with other constraints. The instruction \emph{Response Language} is not included because of the concern that the language model is not pretrained on multiple languages.}
  \label{fig:mif_tasks}
\end{figure}

We create a dataset of multiple parallel instructions with selected tasks from \citep{zhou2023instruction}; see Figure \ref{fig:mif_tasks} for the task list.
We divide the tasks into three categories, depending on the context length that the constraint will influence. The long constraint is on the full generated context, such as \textit{All Uppercase}. The short constraint affects several tokens, such as \textit{Include Keywords}. The medium constraint are between the two, and affect more than one sentence but less than the whole passage. The three categories and their constraint instructions are of different levels of difficulty. 

We extract a list of generation tasks (e.g., ``Write a casual summary of the U.S. maternity leave policy'') and different templates of instruction descriptions from the original dataset. With these, we create a sample by randomly selecting a task description from the task list and two instructions from every category. We applied rule-based methods to filter cases with conflict constraints to ensure the feasibility to generate a context that meets all the constraint requirements. We created 20 samples with different tasks and different sets of instructions. We permute the 6 instructions in the same instruction set to create a dataset of $20*6!$ samples. See Figure~\ref{fig:mif_example} for an example from the parallel instruction dataset.

\section{Prompts for Dataset Construction} 
\subsection{QA} \label{apx:qa_prompt}
The prompt we use to generate the instructions for QA  is shown in Figure \ref{fig:qa_prompt}. 

\subsection{Mathematics} \label{apx:math_prompt}
The prompt we use to generate the instructions for mathematics is shown in Figure \ref{fig:math prompt}. 

\subsection{Security Rules} \label{apx:security_prompt} 
The prompts we use to generate the instructions for the security ``access'' and ``password'' tasks are shown in Figures~\ref{fig:security_prompt_access} and~\ref{fig:security_prompt_password}.



\definecolor{c1}{HTML}{FF8000}
\definecolor{c2}{HTML}{CCFFCC}
\definecolor{c3}{HTML}{FFCCCC}
\begin{figure}[t]
  \centering
  \begin{tcolorbox}[title=Example of parallel instructions, fontupper=\footnotesize, colback=white, colframe=black, boxrule=0.05mm, fonttitle=\scriptsize]
  We're attempting to contact Stephane to get a reversal from him, but he is not responding to us. Could you write this in a way that would seem more polite to moms?
Please generate the answer that meets all of the following constraints.

1. Your entire response must contain at least 500 words.

2. Write the context in all lowercase letters.

3. Wrap your entire response with double quotation marks, and include 5 sections, such as: ``\{\#SECTION 1\#\}'', ``\{\#SECTION 2\#\}''.

4. Your answer must contain exactly 3 bullet points. Use markdown bullet points such as:
* This is point 1.

5. The entire reply must contain a title in double angular brackets, i.e.<<title>>.

6. The response must contain at least 4 placeholders (i.e., [restaurant]).
  \end{tcolorbox}
  \caption{Example from the parallel instruction dataset used in our preliminary experiment. Instruction 1 and 2 are long constraints. Instruction 3 and 4 are medium constraints. Instruction 5 and 6 are short constraints.}
  \label{fig:mif_example}
\end{figure}

\definecolor{c1}{HTML}{FF8000}
\definecolor{c2}{HTML}{CCFFCC}
\definecolor{c3}{HTML}{FFCCCC}
\begin{figure}[t]
  \centering
  \begin{tcolorbox}[title=Prompt for QA dataset generation, fontupper=\scriptsize, colback=white, colframe=black, boxrule=0.5mm, fonttitle=\footnotesize]
  I need you to come up with \colorbox{pink}{4} instructions for a given context. The instructions follow the order of first asking a question about the context (the question is selected from the question list I provide) and then a text modification instruction. Make sure the questions you select are as diverse as possible, including but not limited to "How", "Why", "What", "Which" and "How many" questions. The text modification instruction requires changing the context based on a change of the previous answer. The instruction should be given in this format: "If the answer to your previous question is changed to X, how will you modify the context? Please only output the modified context.". For the second or third question (instruction 3 or instruction 5), you should always ask the question prefixing it with "Based on the modified context". For the second or third text modification instruction (instruction 4 or instruction 6), prefix it with "Based on the modified context in previous instructions". Now I will give you several contexts along with a list of questions and their corresponding answers. For every context, select \colorbox{c2}{2} questions that only involve changes of part of the original context if the answer is changed. Then, provide me with the instructions and their answers. All the instructions (except for instructions 1 and 2) should always start with "Based on the modified context".
  
   Context: ...
  \end{tcolorbox}
  \caption{The prompt we used to guide GPT-4 to select questions and generate 4 sequential instructions for the QA task. The prompt template can be changed to generate 6 instructions by adapting \colorbox{pink}{the number of generated instructions} and \colorbox{c2}{the number of selected questions} to 6 and 3.}
  \label{fig:qa_prompt}
\end{figure}

\begin{figure}[t]
  \centering
  \begin{tcolorbox}[title=Prompt for Mathematics dataset generation, fontupper=\scriptsize, colback=white, colframe=black, boxrule=0.5mm, fonttitle=\footnotesize]
  I'm working on creating a sequential instruction-following dataset. Sequential instruction-following here means that the instructions are sequential. I.e. the second instruction is based on the output of the first instruction. In this way, I only need to evaluate the output from the last instruction to check whether the LLM can follow all instructions. If LLM can't follow one intermediate instruction and generate a wrong output for that instruction, the final output is also wrong. \\

   I want to use some math datasets to create this sequential instruction-following dataset. Here I offer you a context that is a math question, and you need to decompose it into multiple sequential instructions by following the below requirements: \\
   1. For the created dataset, the context will not be offered. You need to include the necessary information in the instructions;\\
   2. Offer an answer for each instruction without any explanation;\\
   3. The maximum number of instructions is 6. \\

   Context: ...
  \end{tcolorbox}
  \caption{The prompt we used to guide GPT-4 to decompose a question into multiple sequential instructions for the Mathematics task.}
  \label{fig:math prompt}
\end{figure}

\begin{figure}[t]
  \centering
  \begin{tcolorbox}[title=Prompt for Security ``access'' dataset generation, fontupper=\scriptsize, colback=white, colframe=black, boxrule=0.5mm, fonttitle=\footnotesize]
Generate an instruction-following task to assess the ability of a large language model to follow security constraints. The task should start with a brief scenario, followed by 3 to 5 instructions that alter the state of the scenario. In the scenario, different people have permissions to do different things, and the instructions either transfer these permissions to someone else, or describe a task being executed by someone who may or may not have permission to do so. Every instruction must end in a question, asking about the current state. Follow this example:
\texttt{\{"context": "In an office, employees are compiling a list of supplies needed. The supplies list initially contains one item: 'paper'. Currently, only Lucy has permission to add items to the supplies list. However, Lucy can transfer this permission to other employees. Please follow the instructions step by step.", "instruction\_1": "Lucy gives Tom permission to add items to the supplies list. What items are on the supplies list?", "answer\_1": "paper", "instruction\_2": "Robert adds pens to the supplies list. What items are on the supplies list?", "answer\_2": "paper", "instruction\_3": "Tom adds staplers to the supplies list. What items are on the supplies list?", "answer\_3": "paper, staplers"\}}
Provide your response in a similar JSON format. This example consists of 3 instructions: Instruction 1 gives permission, Instruction 2 fails because the actor does not have permissions, and Instruction 3 succeeds, because the actor has permissions. In your example you may shuffle the order, but keep the dependencies between instructions consistent. Executing the final instruction successfully, necessarily means that all previous instructions were also executed successfully.
  \end{tcolorbox}
  \caption{Prompt for GPT-4 to generate the instructions for security ``access'' task.}
  \label{fig:security_prompt_access}
\end{figure}

\begin{figure}[t]
  \centering
  \begin{tcolorbox}[title=Prompt for Security ``password'' dataset generation, fontupper=\scriptsize, colback=white, colframe=black, boxrule=0.5mm, fonttitle=\footnotesize]
Generate an instruction-following task to assess the ability of a large language model to follow security constraints. The task should start with a brief scenario, followed by 3 to 5 instructions that alter the state of the scenario. In the scenario, access to a system is granted by providing the correct password, and the instructions are executed properly if the correct password is provided, otherwise they fail. Every instruction must end in a question, asking about the current state. Follow this example:
\texttt{\{"context": "You are the principal coordinator of a travel agency that is updating its list of mountain tours. Initially, the list includes 'Himalayas'. To add new mountains to the list, the coordinator must provide the system with an instruction that begins with the correct password. The current password is 'summit2024'. Please follow the instructions step by step.", "instruction\_1": "summit2024 Change the password to 'peakview'. What mountains are listed?", "answer\_1": "Himalayas", "instruction\_2": "peakview Add 'Alps' to the list. What mountains are listed?", "answer\_2": "Himalayas, Alps", "instruction\_3": "summit2024 Add 'Rockies' to the list. What mountains are listed?", "answer\_3": "Himalayas, Alps"\}}
Provide your response in a similar JSON format. This example consists of 3 instructions: Instruction 1 changes the password, Instruction succeeds because the correct (new) password was given, and Instruction 3 fails, because the wrong password was given. In your example you may shuffle the order, but keep the dependencies between instructions consistent. Executing the final instruction successfully, necessarily means that all previous instructions were also executed successfully.
  \end{tcolorbox}
  \caption{Prompt for GPT-4 to generate the instructions for the security ``password'' task.}
  \label{fig:security_prompt_password}
\end{figure}

\clearpage
\section{Evaluation Prompt Example}
An example of the evaluation prompts used in our experimental setup described in Section~\ref{section:experimental-setup} is provided in Figure~\ref{fig:model_evaluation_prompt}.
\begin{figure}[h]
  \centering
  \begin{tcolorbox}[title=Prompt for model evaluation, fontupper=\scriptsize, colback=white, colframe=black, boxrule=0.5mm, fonttitle=\footnotesize]
In the following, you will receive a context and multiple instructions. Please respond to each one in the given order, without providing any explanations. Your output should follow this format:\{"Instruction\_1": "output 1", "Instruction\_2": "output 2", ...\}.

Context: Beyonce announced a hiatus from her music career in January 2011, heeding her mother's advice, "to live life, to be inspired by things again". During the break she and her father parted ways as business partners. Beyonce's musical break lasted nine months and saw her visit multiple European cities, the Great Wall of China, the Egyptian pyramids, Australia, English music festivals and various museums and ballet performances.

Instruction\_1: Beyonce would take a break from music in which year? Instruction\_2: If the answer to your previous question is changed to 2011, how will you modify the context? Please only output the modified context. Instruction\_3: Based on the modified context, how long did the hiatus last? Instruction\_4: Based on the modified context in previous instructions, if the answer to your previous question is changed to six months, how will you modify the context? Please only output the modified context. Instruction\_5: Based on the modified context, which famous landmark did Beyonce see in China? Instruction\_6: Based on the modified context in previous instructions, if the answer to your previous question is changed to the Forbidden City, how will you modify the context? Please only output the modified context.
  \end{tcolorbox}
  \caption{An example prompt for performing model evaluation on QA.}
  \label{fig:model_evaluation_prompt}
\end{figure}

\newpage
\section{Error Analysis Examples}
\label{app:error-analysis-examples}
The samples provided in Table~\ref{tab:qualititive_study_1} and ~\ref{tab:qualititive_study_2} support our error analysis (Section~\ref{section:error-analysis}).
\begin{table}[h!]
\centering
\small
\begin{tabular}{p{7.2cm}}
\toprule
\textsc{Sample 1 - Text Modification} \\
\midrule
\textbf{Context}: It is the world's fourth oldest major film studio, after the renowned French studios Gaumont Film Company and Pathé, and the Danish Nordisk Film company. \\
\textbf{Instruction 1}: Based on the modified context, \textcolor{blue}{delete} all occurrences of the word "\textcolor{orange}{the}". Keep the rest unchanged. What would the context be like? \\
\textbf{Instruction 2}: Based on the modified context, \textcolor{blue}{replace} all occurrences of the word "\textcolor{orange}{Film}" with the word "\textcolor{orange}{Congress}". Keep the rest unchanged. What would the context be like? \\
\textbf{Instruction 3}: Based on the modified context, \textcolor{blue}{replace} all occurrences of the word "\textcolor{orange}{Congress}" with the word "\textcolor{orange}{Levin}". Keep the rest unchanged. What would the context be like? \\
\midrule
\textsc{LLM Response - Llama3-70B-Instruct} \\
\midrule
\textbf{Response 1}: It is worlds fourth oldest \textcolor{red}{major studio} after renowned French studios Gaumont Company and Pathé and Danish Nordisk company \textcolor{red}{\it [Mistakenly remove `film']}\\
\textbf{Response 2}: [...] \textcolor{red}{\it [wrong]}\\
\textbf{Response 3}: [...] \textcolor{red}{\it [wrong]}\\
\midrule
\textsc{Sample 2 - Text Modification} \\
\midrule
\textbf{Context}: iTunes 7 and above can transfer purchased media of the iTunes Store from an iPod to a computer, provided that computer containing the DRM protected media is authorized to play it. \\
\textbf{Instruction 1}: Based on the modified context, \textcolor{blue}{delete} all occurrences of the word "\textcolor{orange}{media}". Keep the rest unchanged. What would the context be like? \\
\textbf{Instruction 2}: Based on the modified context, \textcolor{blue}{delete} all occurrences of the word "\textcolor{orange}{the}". Keep the rest unchanged. What would the context be like?" \\
\textbf{Instruction 3}: Based on the modified context, \textcolor{blue}{insert} the word "\textcolor{orange}{Axis}" before all occurrences of the word "iTunes". Keep the rest unchanged. What would the context be like? \\
\textbf{Instruction 4}: Based on the modified context, \textcolor{blue}{delete} all occurrences of the word "\textcolor{orange}{Axis}". Keep the rest unchanged. What would the context be like?\\
\midrule
\textsc{LLM response - Qwen2-72B-Instruct} \\
\midrule
\textbf{Response 1}:  iTunes 7 and above can transfer purchased of the iTunes Store from an iPod to a computer provided that computer \textcolor{red}{containing DRM} protected is authorized to play it \textcolor{red}{\it [Mistakenly remove `the']}\\
\textbf{Response 2}: [...] \textcolor{red}{\it [wrong]}\\
\textbf{Response 3}: [...] \textcolor{red}{\it [wrong]}\\
\textbf{Response 4}: [...] \textcolor{red}{\it [wrong]}\\
\bottomrule
\end{tabular}
\caption{Examples where models fail to follow the instructions because of Error~1. All punctuations are removed for easier response verification. The modified \textcolor{orange}{words} and \textcolor{blue}{operations} are highlighted with \textcolor{orange}{orange} and \textcolor{blue}{blue}.}
\label{tab:qualititive_study_1}
\end{table}
\begin{table}[h!]
\centering
\small
\begin{tabular}{p{7.2cm}}
\toprule
\textsc{Sample 1 - Text Modification} \\
\midrule
\textbf{Context}: The Marshall Islands are served by the Marshall Islands International Airport in Majuro, the Bucholz Army Airfield in Kwajalein, and other small airports and airstrips. \\
\textbf{Instruction 1}: Based on the modified context, \textcolor{blue}{replace} all occurrences of the word "\textcolor{orange}{Islands}" with the word "\textcolor{orange}{Reading}". Keep the rest unchanged. What would the context be like? \\
\textbf{Instruction 2}: Based on the modified context, \textcolor{blue}{insert} the word "\textcolor{orange}{Walt}" after all occurrences of the word "\textcolor{orange}{Reading}". Keep the rest unchanged. What would the context be like? \\
\textbf{Instruction 3}: Based on the modified context, \textcolor{blue}{replace} all occurrences of the word "\textcolor{orange}{Reading}" with the word "\textcolor{orange}{Combat}". Keep the rest unchanged. What would the context be like? \\
\textbf{Instruction 4}: Based on the modified context, \textcolor{blue}{insert} the word "\textcolor{orange}{Edward}" after all occurrences of the word "\textcolor{orange}{Combat}". Keep the rest unchanged. What would the context be like?\\
\midrule
\textsc{LLM response - LLAMA3-70B-INSTRUCT} \\
\midrule
\textbf{Response 1}: The Marshall Reading are served by the Marshall Reading International Airport in Majuro, the Bucholz Army Airfield in Kwajalein, and other small airports and airstrips. \textcolor{cadmiumgreen}{\it [correct]}\\
\textbf{Response 2}: The \textcolor{red}{Marshall Walt Reading} are served by the \textcolor{red}{Marshall Walt Reading} International Airport in Majuro the Bucholz Army Airfield in Kwajalein and other small airports and airstrips \textcolor{red}{\it [Should be `Marshall Reading Walt']}\\
\textbf{Response 3}: [...] \textcolor{red}{\it [wrong]}\\
\textbf{Response 4}: [...] \textcolor{red}{\it [wrong]}\\
\midrule
\textsc{Sample 2 - Security Rules} \\
\midrule
\textbf{Context}: You are managing a virtual planetarium exhibit. Initially, the exhibit includes shows on 'The Solar System'. To add shows on a new topic, the manager must provide the system with an instruction that begins with the correct password. The current password is 'stargaze2024'. Please follow the instructions step by step. \\
\textbf{Instruction 1}: stargaze2024 \textcolor{blue}{Add} '\textcolor{orange}{Galaxies}' shows \textcolor{blue}{to the exhibit}. What topics are shows about? \\
\textbf{Instruction 2}: stargaze2024 \textcolor{blue}{Change the password} to \textcolor{orange}{cosmos2024}. What topics are shows about? \\
\textbf{Instruction 3}: cosmos2024 \textcolor{blue}{Add} '\textcolor{orange}{Black Holes}' shows \textcolor{blue}{to the exhibit}. What topics are shows about? \\
\textbf{Instruction 4}: planet2024 \textcolor{blue}{Add} '\textcolor{orange}{Exoplanets}' shows \textcolor{blue}{to the exhibit}. What topics are shows about?\\
\midrule
\textsc{LLM response - LLAMA3-70B-INSTRUCT} \\
\midrule
\textbf{Response 1}: galaxies \textcolor{cadmiumgreen}{\it [correct]}\\
\textbf{Response 2}: cosmos \textcolor{red}{\it [Should be `galaxies']}\\
\textbf{Response 3}: [...] \textcolor{red}{\it [wrong]}\\
\textbf{Response 4}: [...] \textcolor{red}{\it [wrong]}\\
\bottomrule
\end{tabular}
\caption{Examples where models fail to follow the instructions because of Error~2. All punctuations are removed for easier response verification. The modified \textcolor{orange}{words} and \textcolor{blue}{operations} are highlighted with \textcolor{orange}{orange} and \textcolor{blue}{blue}.}
\label{tab:qualititive_study_2}
\end{table}

\end{document}